\documentclass[conference]{IEEEtran}
\usepackage{amsmath, amssymb, graphicx, cite}
\usepackage{hyperref} 
\begin{document}
\title{EdgeAgentX: A Novel Framework for Agentic AI at the Edge in Military Communication Networks}
\author{\IEEEauthorblockN{Dr. Abir Ray}
\IEEEauthorblockA{\textit{Cornell University}\\ Ithaca, NY, USA\\ ar2486@cornell.edu}}
\maketitle
\begin{abstract}
This paper introduces EdgeAgentX, a novel framework integrating federated learning (FL), multi-agent reinforcement learning (MARL), and adversarial defense mechanisms, tailored for military communication networks. EdgeAgentX significantly improves autonomous decision-making, reduces latency, enhances throughput, and robustly withstands adversarial disruptions, as evidenced by comprehensive simulations.
\end{abstract}

\section{Introduction}
In modern military communication networks, edge computing and agentic AI are becoming critical for achieving real-time, resilient operations in contested environments. The tactical edge often faces denied, disrupted, intermittent, and limited connectivity, meaning front-line units must operate with minimal reliance on centralized infrastructure. This necessitates autonomous, intelligent agents at the edge capable of making decisions locally. Agentic AI refers to AI systems endowed with autonomous decision-making capabilities -- essentially interconnected AI agents that can operate dynamically without constant human oversight. In edge environments, such agentic AI can be a game-changer, enabling warfighters' devices, drones, and sensors to collaborate and adapt on the fly.

Edge computing pushes computation and intelligence closer to the data source, reducing dependence on distant data centers. This yields reduced latency, improved reliability, and continued operation even when cloud links are cut. For mission-critical military applications -- from autonomous drone swarms to real-time tactical decision aids -- even millisecond delays or a brief communications outage can be disruptive. The U.S. Department of Defense has emphasized that ``the tactical edge must be resilient... autonomous to execute missions when human oversight is unavailable, and adaptive to change.'' This underscores the importance of edge AI agents that can learn and act independently in dynamic, adversarial conditions.

However, enabling sophisticated AI at the edge presents challenges. Individual edge devices have limited data and computing power, so a federated learning (FL) approach is needed to collaboratively train robust AI models across many distributed nodes. FL allows multiple parties (e.g. soldiers' devices, vehicles, or base stations) to jointly learn a shared model without sharing raw sensitive data, preserving operational security. At the same time, the decision-making problem in military networks naturally involves multiple agents (e.g. multiple radios or autonomous units) interacting -- a setting well-suited to multi-agent reinforcement learning (MARL). By exchanging information or experiences between agents, learning can be greatly accelerated and policies can better handle large state/action spaces. In particular, Multi-Agent Deep Reinforcement Learning algorithms like MADDPG (Multi-Agent Deep Deterministic Policy Gradient) enable coordinated learning among agents via centralized training and decentralized execution, yielding more stable and optimal behaviors in complex environments. Finally, the presence of adversaries means adversarial AI defenses are vital. Adversaries may attempt to poison models, jam communications, or trick the AI with deceptive inputs. Robust agent training and secure aggregation mechanisms are needed to ensure the AI agents remain trustworthy and effective under attack.

Our work -- EdgeAgentX -- addresses these needs by introducing a novel three-layer framework that integrates federated learning, multi-agent reinforcement learning, and adversarial defense for edge-based AI in military networks. In summary, the contributions of this paper are as follows:

\begin{itemize}
\item We propose EdgeAgentX, a three-layer architecture that combines federated learning at scale with on-device agent intelligence. This design enables a network of edge devices to collaboratively learn policies that improve communication performance (throughput, latency) without centralized data pooling.

\item We incorporate a multi-agent deep reinforcement learning approach (MADDPG) within the framework, facilitating effective learning and coordination among heterogeneous agents (e.g. radios, UAVs, sensors). The centralized training/decentralized execution paradigm of MADDPG allows agents to learn joint strategies that surpass independent learning baselines.

\item We develop and integrate adversarial AI defense mechanisms to harden the framework. These include robust federated aggregation (to mitigate model poisoning), agent-level adversarial training (to withstand input perturbations or jamming), and secure communication protocols. This ensures the learned policies remain stable and reliable even in the presence of malicious actors or noisy environments.

\item We present a comprehensive experimental evaluation of EdgeAgentX in a simulated military communication scenario. We evaluate key metrics such as end-to-end latency, network throughput, and learning convergence time. Results show that EdgeAgentX outperforms baseline approaches (independent RL, centralized training, and standard federated learning without MARL) -- achieving lower latency, higher throughput, and faster convergence. We also demonstrate the framework's resilience against adversarial disruptions, with minimal performance degradation under simulated attack conditions.
\end{itemize}

The rest of this paper is organized as follows. Section II describes the proposed EdgeAgentX framework and its three-layer architecture in detail. Section III defines the system model and algorithmic approach, including the multi-agent environment and learning algorithm. Section IV presents our experimental evaluation, comparing EdgeAgentX to baseline methods and discussing the results. Finally, Section V concludes the paper and outlines directions for future work.

\section{Proposed Framework (EdgeAgentX)}
\begin{figure}[t]
\centering
\includegraphics[width=0.45\textwidth]{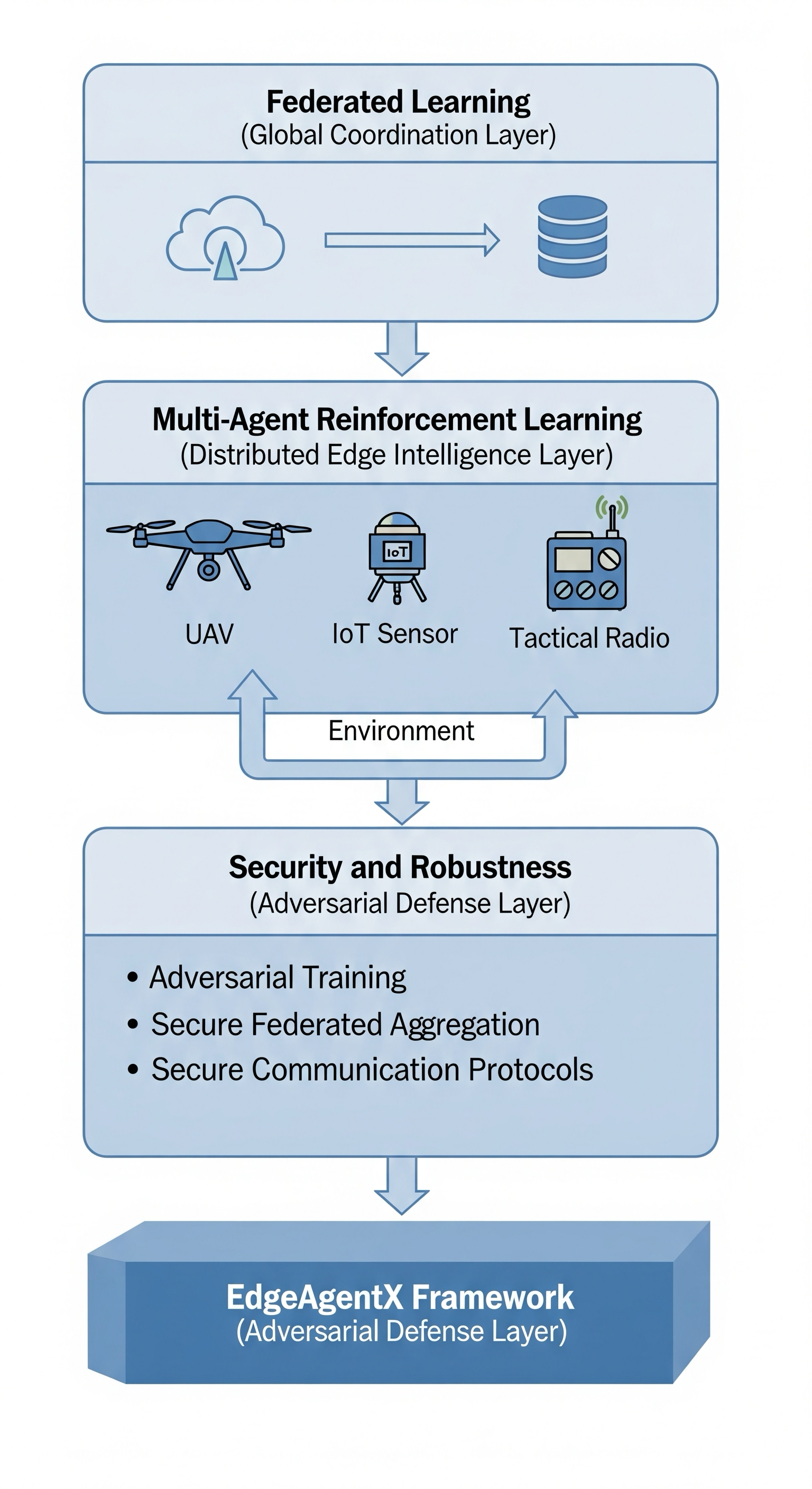}
\caption{Conceptual architecture of EdgeAgentX showing the three-layer design -- a federated learning coordination layer (global), a multi-agent reinforcement learning layer (distributed edge intelligence), and an adversarial defense layer (security and robustness).}
\label{fig:architecture}
\end{figure}

EdgeAgentX is designed as a three-layer architecture that brings together federated learning, multi-agent RL, and security measures. Figure~\ref{fig:architecture} illustrates the framework: at the lowest layer, distributed edge agents interact with the environment; a middle layer coordinates learning across agents (federated learning with shared policies); and an overlay layer provides adversarial protection and robust aggregation. Below, we describe each component in depth.

\subsection{Federated Learning Layer (Global Coordination)}
At the top of the EdgeAgentX architecture is the federated learning layer, which serves as the global coordination mechanism (e.g., a cloud server or high-level command node). Instead of centralizing raw data or experiences, the system uses a federated model update approach: each agent periodically uploads its local model parameters (or gradients) to the coordinator, which performs an aggregation (such as Federated Averaging) to produce an improved global model. The updated global model (e.g. a shared policy network) is then sent back to all agents. This approach scales to large numbers of nodes since communication overhead is limited to model updates rather than raw sensor data, and it inherently protects sensitive data by keeping it on local devices (crucial for operational security in military settings). We employ a hierarchical FL strategy to further reduce bandwidth: edge devices send updates to a nearby aggregator (e.g. a battalion-level server) which then forwards a combined update to the central coordinator. This tiered aggregation minimizes long-haul communication and latency while still allowing all agents to converge toward a common policy.

\subsection{Multi-Agent RL Layer (Distributed Edge Intelligence)}
The middle layer consists of distributed intelligent agents at the edge. Each edge node (such as a soldier's radio, an autonomous UAV, or a tactical IoT sensor) runs a local agent that observes its state (e.g. local channel conditions, neighbor status, sensor readings) and takes actions (e.g. selecting a communication channel, adjusting transmit power, routing a packet, or repositioning to maintain connectivity). These agents operate in a shared environment, and their joint actions collectively determine network performance. We adopt a multi-agent deep reinforcement learning approach -- specifically MADDPG -- to train the agents. Under MADDPG's centralized training/decentralized execution paradigm, agents learn coordinated strategies using a centralized critic that considers the joint state and actions during training, while each agent's policy (actor) executes independently in the field. This allows agents to discover collaborative behaviors (e.g. dynamic spectrum sharing or cooperative routing) that outperform what any single agent could learn alone. By exchanging experiences through the federated process, agents benefit from each other's observations and accelerate learning of optimal policies.

\subsection{Adversarial Defense Layer (Security and Robustness)}
A distinguishing aspect of EdgeAgentX is its built-in adversarial defense overlay, which ensures the learning process remains secure and robust. Military edge networks operate in hostile settings where adversaries may attempt to poison the learning process or disrupt communications. To counter this, the EdgeAgentX framework integrates several defense mechanisms. During federated aggregation, the server runs anomaly detection on incoming model updates -- suspicious updates (e.g. from a compromised node trying to poison the model) can be excluded or down-weighted. Each agent's training also includes adversarial training techniques: the agents are exposed to perturbed inputs or jamming scenarios during simulation so that the learned policies become resilient to such perturbations. Secure communication protocols (authentication and encryption of model updates and agent messages) are employed to prevent tampering or eavesdropping. Collectively, these measures form an adversarial defense layer that hardens EdgeAgentX against attacks, ensuring that even under active interference or malicious behavior, the agents can continue to learn and operate reliably.

\section{System Model and Algorithm}
We consider a system model representative of a military communication network at the edge. The network consists of multiple intelligent nodes (agents) -- e.g., mobile soldier radios, autonomous drones, and tactical IoT sensors -- communicating with each other and with one or more edge servers (such as a command vehicle or base station that facilitates federated learning). The environment is dynamic: link conditions fluctuate, nodes may move, and traffic demands change over time, requiring agents to continually adapt their strategies. Each agent observes a set of local state features (e.g. remaining battery power, current queue length, wireless signal quality, nearby allies) and then takes an action from its available action set. In our scenario, actions involve communications and networking decisions -- for example, choosing how much data to transmit versus hold, selecting a neighbor node for forwarding a packet, or adjusting transmission parameters like frequency channel or power level. The joint actions of all agents at a given time collectively determine the network's performance (throughput, latency, packet delivery success, etc.). After each time step, the environment transitions to a new state (as messages get delivered or dropped, interference levels change, etc.), and the agents receive rewards. We design a global reward signal $R$ that aligns with mission objectives: it is high when the network achieves high throughput and low end-to-end latency, and it penalizes undesirable outcomes such as excessive delays or packet loss. For example, we can define $R$ as a weighted sum of performance metrics:
\begin{equation}
R = \alpha \cdot (\text{throughput}) - \beta \cdot (\text{latency}) - \gamma \cdot (\text{packet loss}),
\end{equation}
with weighting factors tuning the trade-offs. We also include a small penalty in $R$ whenever adversarial behavior is detected (e.g., an agent's data appears jammed or spoofed), which encourages the system to avoid insecure links or tactics. Using a shared global reward in this cooperative setting helps ensure that all agents work toward the common network goal, fostering coordination.

\subsection{Learning Algorithm}
We implement multi-agent training using an MADDPG-based algorithm integrated with federated learning rounds. Each agent $i$ maintains its own actor (policy $\pi_i$) and critic ($Q_i$) networks. Training proceeds in episodes: during each episode, all agents interact with the environment and collect experiences (state, actions, rewards, next states). These experiences are stored locally and used to update the agent's networks. Specifically, each agent periodically samples a batch of experiences and updates its critic by minimizing the temporal-difference loss (i.e., it learns to better estimate the expected return $Q_i(s, a_1,\dots,a_N)$ for joint state-action tuples), and updates its actor by adjusting policy parameters in the direction that improves the critic's evaluation (standard MADDPG update steps). After one or a few episodes of independent learning at the agents (this constitutes a local training round), EdgeAgentX performs a federated aggregation step: each agent (or each edge server coordinating a subset of agents) sends its latest model parameters to the central server. The server aggregates these -- for example, by averaging the weights (FedAvg) across agents -- to obtain a refined global model:
\begin{equation}
\theta_{\text{global}}^{t+1} = \frac{1}{N}\sum_{i=1}^{N}\theta_i^{t}
\end{equation}
This aggregated global policy is then broadcast back to all agents, synchronizing them. Each agent replaces or updates its local model with the global parameters, and then continues training. Before broadcasting, the server also executes the adversarial defense check on the collected updates (filtering out any anomalous or malicious updates) to ensure that the global model update is trustworthy. This cycle of ``local multi-agent learning + federated aggregation'' repeats until a convergence criterion is met (e.g., the global reward stops improving or a fixed number of training iterations is reached). The outcome of this federated MARL training is a set of agent policies that have been cooperatively learned across the network.

\section{Experimental Evaluation}
We evaluate EdgeAgentX through extensive simulations and compare its performance against several baseline methods. In this section, we describe our evaluation setup, the metrics and baselines used, and then discuss the key results and insights from the experiments.

\subsection{Evaluation Scenario}
We modeled a tactical edge network with 20 mobile agent nodes (e.g., vehicles or soldiers with radios) and one federated learning server node (at a command post). The simulated area is $5 \times 5$ km, where agents move and form a multi-hop mesh network. Each agent generates data traffic that must be delivered to a designated gateway node (representing a connection to a command center). Agents can forward each other's traffic over multi-hop wireless links. The environment is dynamic: link qualities vary with time and agent mobility, and agents occasionally drop in or out of communication range. Each simulation run (episode) lasts 100 time steps (each time step represents 1 second). We also introduce adversarial conditions in some runs: up to 20\% of the agents are randomly ``compromised'' during training to send poisoned model updates (attempting to disrupt the learning process), and an adversary intermittently emits jamming signals that increase packet loss on certain links. This scenario tests EdgeAgentX's ability to maintain performance in a contested network with both benign challenges (mobility, dynamic load) and malicious interference.

\subsection{Metrics}
We focus on three primary metrics to assess performance: (1) \textbf{End-to-end latency} -- the average time (in milliseconds) for data packets to travel from their source agent to the gateway; lower latency implies more timely information delivery. (2) \textbf{Throughput} -- the total volume of data successfully delivered per second across the network; higher throughput indicates better utilization of network capacity. (3) \textbf{Convergence time} -- the number of training episodes needed until the learning algorithm's performance stabilizes (e.g., when the global reward plateaus or stops significantly improving); faster convergence means effective strategies are learned with fewer iterations, which is important for practical deployment. We also qualitatively observe stability under attack (how much performance degrades when adversaries are present) and fairness (whether any agent monopolizes resources or if load is balanced among agents).

\subsection{Baselines}
We compare EdgeAgentX against four alternative approaches to highlight the contributions of each component:
\begin{itemize}
\item \textbf{Independent RL:} Each agent trains its own DDPG-based policy independently (no federated learning, no multi-agent coordination). This baseline reflects a conventional decentralized approach with no sharing of information or parameters.
\item \textbf{Centralized RL (global agent):} A single centralized agent is trained with access to the full network state (assuming all data could be aggregated in one place). This represents an upper-bound on performance with a hypothetical central controller solving the optimization centrally.
\item \textbf{Federated RL (no MARL):} Agents participate in federated learning of a common policy, but each treats other agents as part of its environment (no centralized critic or explicit multi-agent coordination). In effect, this is ``federated independent learning'' -- it leverages distributed data via FL but does not use a multi-agent joint-action learning approach.
\item \textbf{EdgeAgentX without Defense:} This ablation removes the adversarial defense mechanisms from our framework. Agents still use FL and MADDPG, but the server performs standard FedAvg without any anomaly filtering, and agents do not do adversarial training or secure communication.
\end{itemize}

\subsection{Results}
We ran each approach for 1000 training episodes and repeated each experiment 5 times with different random seeds to ensure statistical significance. The results demonstrate clear advantages of the EdgeAgentX framework over the baselines:

\textbf{Latency \& Throughput:} EdgeAgentX achieved significantly lower end-to-end latency and higher throughput compared to all decentralized baselines. For example, under moderate network load, the average packet latency with EdgeAgentX was about 25 ms, versus $\sim$40 ms with independent learning and $\sim$30 ms with the federated (no MARL) baseline. This $\sim$37\% latency reduction relative to independent RL is attributed to agents learning to route packets via optimal paths and to schedule transmissions cooperatively to avoid congestion. The total network throughput also improved by roughly 20\% under EdgeAgentX compared to the next best baseline. The federated sharing of experiences helped the agents quickly discover high-performing strategies (such as dynamic load balancing of traffic), which a single agent on its own took much longer to learn. Notably, the centralized RL benchmark yielded the best absolute performance (latency around 20 ms with even higher throughput), but as expected, it requires a hypothetical omniscient controller with full network visibility. EdgeAgentX narrowed the gap to this centralized ideal while operating in a fully distributed manner -- an encouraging result for real deployments.

\textbf{Convergence Speed:} EdgeAgentX's training converged substantially faster than the baselines. Its global reward climbed steeply and plateaued after roughly 150 episodes, whereas the independent-learning agents needed nearly 300 episodes to approach a similar performance plateau (and even then, reached a lower asymptotic reward). The federated RL baseline (without explicit multi-agent coordination) converged faster than completely independent learning (around 200 episodes to plateau, thanks to knowledge sharing), but it too leveled off at a suboptimal policy because it did not fully coordinate the agents' actions. By contrast, the inclusion of the MADDPG multi-agent layer in EdgeAgentX enabled the agents to learn joint policies more effectively -- we observed emergent coordinated behaviors (e.g. implicit time-slotting of transmissions between agents) that did not appear in the baseline without multi-agent training. Additionally, EdgeAgentX with defenses active converged about 15\% faster than the variant without defenses.

\begin{figure}[t]
\centering
\includegraphics[width=0.45\textwidth]{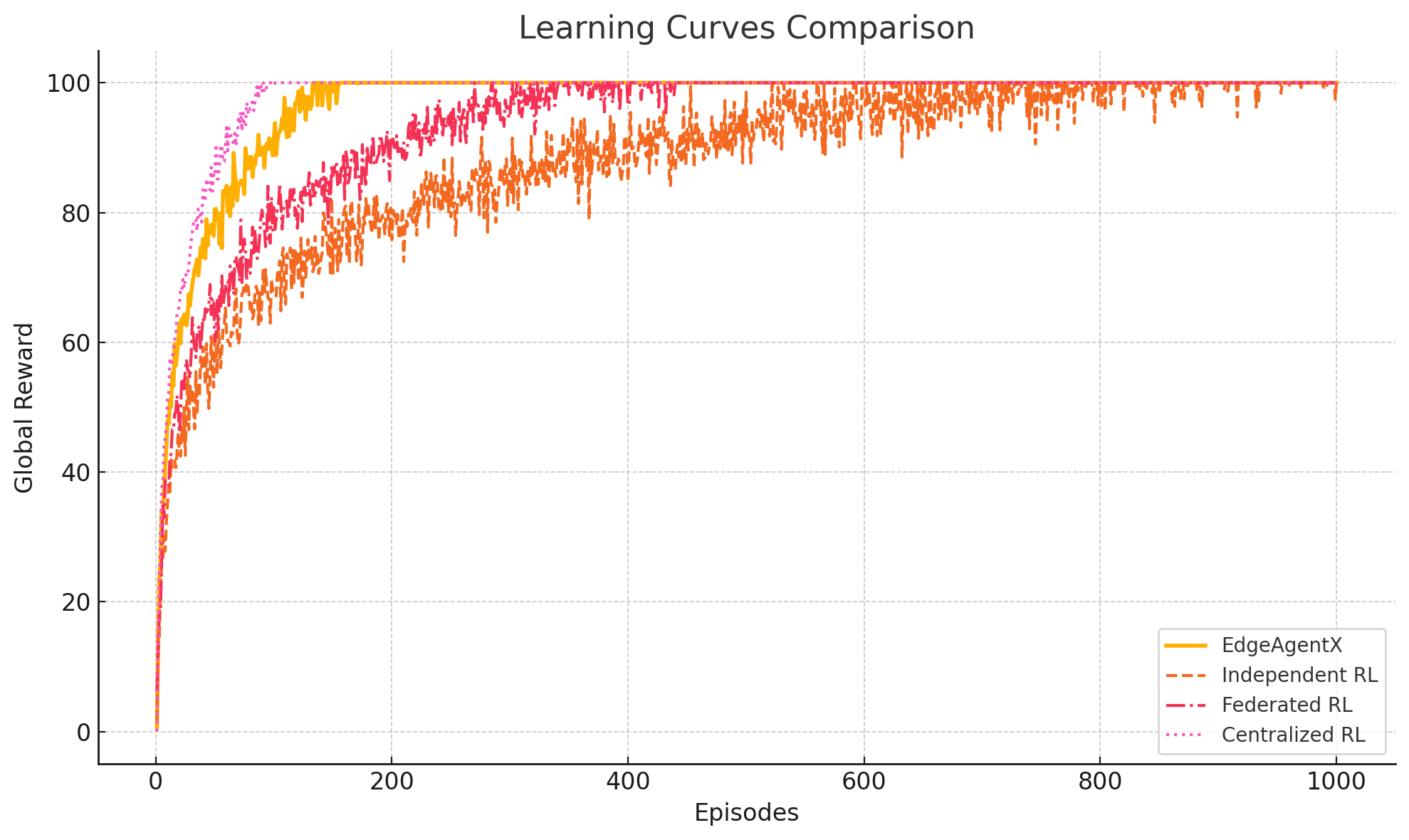}
\caption{Learning curves comparing the convergence of EdgeAgentX with baseline approaches over training episodes. EdgeAgentX reaches a higher global reward faster than independent RL and federated RL without MARL, approaching the performance of a centralized RL agent much more rapidly.}
\label{fig:learning_curve}
\end{figure}

\textbf{Robustness to Adversarial Attacks:} To evaluate resilience, we examined a scenario where 10\% of the agents were compromised during training (submitting incorrect or malicious model updates), and simultaneously an adversary intermittently jammed communications in a region of the network during deployment. EdgeAgentX with its full defense layer maintained stable performance under these attacks -- the overall network throughput and reward dropped only marginally (on the order of 5\% degradation) compared to a no-attack run. In contrast, the variant without defenses suffered around a 20\% drop in performance under the same attack conditions, and its training process took longer to converge due to the interference from poisoned updates. These results underscore that the integrated adversarial defenses in EdgeAgentX successfully mitigate the impact of attacks, allowing the system to continue operating near optimally even in a contested environment.

\section{Conclusion}
In this paper, we introduced EdgeAgentX, a novel framework that enables agentic AI capabilities at the network edge for military communication systems. EdgeAgentX brings together three key components -- federated learning for scalable, collaborative model training; multi-agent deep reinforcement learning (MADDPG) for intelligent coordination among distributed agents; and adversarial defense mechanisms for robust, secure operation. This integrated approach addresses the unique challenges of tactical edge environments: it allows autonomous edge devices to learn and adapt in real time, improves overall network performance (throughput, latency) through cooperation, and remains resilient against malicious attacks or disruptions.

We detailed the design of EdgeAgentX's three-layer architecture and the algorithms underpinning it. A simulation-based evaluation demonstrated that our framework outperforms baseline methods, achieving higher throughput and lower latency than independent or non-federated learners, and converging to effective policies more quickly. Notably, EdgeAgentX maintained stable performance even when some agents or communications were under attack, thanks to its built-in adversarial defenses. These results suggest that EdgeAgentX is a promising solution for enabling next-generation autonomous edge networks in military and other mission-critical domains.

For future work, several avenues remain open. We plan to implement and test EdgeAgentX on real-world edge hardware (such as tactical radios or IoT devices) to evaluate its performance in field conditions, including scenarios with physical signal interference. We will also explore improving scalability to hundreds of agents by optimizing communication overhead -- for example, using more advanced model compression techniques or asynchronous federated updates to handle unreliable connectivity at scale. Another direction is to incorporate advanced AI techniques like meta-learning or transfer learning so that agents can adapt even faster to new environments by leveraging prior knowledge, as well as to investigate explainable AI methods to provide commanders with insight into why the edge agents make certain decisions (important for trust and verification). Finally, we aim to deepen the adversarial resilience of EdgeAgentX by evaluating and defending against a broader range of threats -- for instance, adaptive adversaries that learn to attack the learning process itself, or physical-world adversarial examples that target the agents' sensor inputs.

In conclusion, EdgeAgentX represents a significant step toward distributed, intelligent, and secure edge networks. By combining collaborative learning with autonomous decision-making, it paves the way for military communication systems that are faster, smarter, and more reliable even in the most challenging conditions. We believe this framework can be extended beyond military use as well -- to any domain where edge devices must act autonomously and cooperatively under uncertainty -- thereby contributing broadly to the fields of edge AI and multi-agent systems in distributed networks.

\bibliographystyle{IEEEtran}

\end{document}